\documentclass[11pt,a4paper]{article}

\usepackage[]{emnlp2020}
\usepackage{times}
\usepackage{latexsym}
\usepackage{graphicx}
\usepackage{tipa}
\usepackage{adjustbox}
\usepackage{tabularx}

\usepackage{multirow}
\usepackage{booktabs}
\usepackage{comment}
\usepackage{textcomp}
\usepackage{silence}

\usepackage{xcolor}

\usepackage{hyperref}
\hypersetup{breaklinks=true}
\usepackage{ulem} 
\usepackage[show]{boxnotes}
\newcolumntype{H}{>{\setbox0=\hbox\bgroup}c<{\egroup}@{}}
\newcommand{\eat}[1]{}

\iftrue % set to iftrue for camera ready version

  \renewcommand\sout[1]{\xspace}
\fi

\usepackage{microtype}

\aclfinalcopy % Uncomment this line for the final submission
\title{Improving Similar Language Translation With Transfer Learning}

\author{Ife Adebara ~~~~~~~~  Muhammad Abdul-Mageed \\
\normalsize Natural Language Processing Lab  \\
  \normalsize The University of British Columbia\\
      
  \texttt{ \small \{ife.adebara,muhammad.mageed\}@ubc.ca}
  }

\date{}

\begin{document}

\maketitle
\begin{abstract}
We investigate transfer learning based on pre-trained neural machine translation models to translate between (low-resource) similar languages. This work is part of our contribution to the WMT 2021 Similar Languages Translation Shared Task where we submitted models for different language pairs, including French-Bambara, Spanish-Catalan, and Spanish-Portuguese in both directions. Our models for Catalan-Spanish ($82.79$ BLEU) and Portuguese-Spanish ($87.11$ BLEU) rank top 1 in the official shared task evaluation, and we are the only team to submit models for the French-Bambara pairs.

\end{abstract}

\section{Introduction}
%---------
%\note[mam]{How many language pairs did we submit for? We should say something like: ``We submitted systems for x language pairs: French and Bambara, etc. We should say something about our approach, and spell out the results we get for each pair. We should also say that our models score best on xx pairs using yy setting.}
We present the findings from our participation in the WMT 2021 Similar Language Translation shared task 2021, which focused on translation between similar language pairs in low-resource settings. The similar languages task focuses on building machine translation (MT) systems for translation between pairs of similar languages, without English as a pivot language. 

Similarity between languages interacts with MT quality, usually positively~\cite{adebara2020translating}. Languages described as similar usually share certain levels of mutual intelligibility. Depending on the level of closeness, certain languages may share orthography, lexical, syntactic, and/or semantic structures which may facilitate translation between pairs. However, \textbf{(a) scarcity of parallel data} that can be used for training MT models remains a bottleneck. Even high-resource pairs can suffer from \textbf{(b) low-data quality}. That is, available data is not always guaranteed to be actual bitext with target standing as a translation of source. In fact, some open resources such as OPUS~\cite{tiedemann2012parallel, tiedemann2015morphological} can suffer from noise such as when the source and target sentences belong to the same language. In this work, we tackle both \textbf{(a) scarcity} and \textbf{(b) low-data quality}. For \textbf{a}, we use simple knowledge transfer from already trained MT models to the downstream pair. For \textbf{b}, we use a simple procedure of language identification to remove noisy bitext where both the source and target are detected to be the same language or where source or target is identified as a different language from what it is expected to be. 

The models we develop are for Spanish to Catalan (ES-CA), Catalan to Spanish (CA-ES), Spanish to Portuguese (PT-ES), Portuguese to Spanish (PT-ES), French to Bambara (FR-BM), and Bambara to French (BM-FR) language pairs\footnote{All models are available on \url{https://huggingface.co/Ife/ES-PT}, \url{https://huggingface.co/Ife/PT-ES}, \url{https://huggingface.co/Ife/ES-CA},
\url{https://huggingface.co/Ife/CA-ES},
\url{https://huggingface.co/Ife/FR-BM},
\url{https://huggingface.co/Ife/BM-FR}}. Whenever possible, we choose an available MT model trained with the same source and target languages as our pair of interest. In cases where no such a model exists, we pick a model with either the source or the target language as our intended pair (Section~\ref{model}). To show the utility of our transfer learning approach to the problem, we also train on one pair from scratch (which we treat as a \textit{baseline}). 

%We used Langid.py to clean the training sets. Langid.py identified the language each sentence belong to. We then used the language information to remove every pair that was not classified as one or both of the languages in each language pair. 

We experiment with tokenized \textbf{(primary models)} and untokenized \textbf{(contrastive models)} settings and compared the settings with models developed by fine-tuning pre-trained models as well as models trained from scratch. Our experiments show that the tokenized settings perform better than the untokenized settings for all language pairs. The model fine-tuned on top of the pre-trained MT model has higher performance than our baseline model from the first epoch compared with the model trained from scratch (for six epochs). Our models for the CA-ES and PT-ES language pairs \textit{achieve top 1 rank in the offical shared task results}, with $82.96$ and $47.71$ BLEU scores respectively. In addition, we are the only team that submitted for the rest of the language pairs (i.e., ES-PT, FR-BM, and BR-FR). %These models were built with pre-trained models, with tokenized and cleaned data.

The rest of our paper is organized as follows: we discuss related work in Section~\ref{Lit-review}. We describe the data and pre-processing in Section \ref{Data}. Next, we describe the data cleaning process in Section~\ref{clean}. In Section~\ref{model}, we describe the models we developed for this task and we discuss the various experiments we perform. We also describe the architectures of the models we developed. Then we discuss the evaluation criteria in Section~\ref{Evaluation}. Evaluation is done on both the validation and test sets. In section~\ref{effect} we perform error analysis on the output of our models for some language pairs to determine the types of errors the models make. We conclude with discussion of the insights we gained from the shared task in Section~\ref{Conclusion}. 

%, Catalan to Spanish, and English to French for Spanish to Catalan, Catalan to Spanish, Spanish to Portuguese, Portuguese to Spanish, French to Bambara and Bambara to French respectively. We computed the level of mutual intelligibility between the languages selected and the downstream tasks. 

%the influence of the similarity between source and target languages used for pre-training and the downstream task. Specifically, we wanted to investigate whether similar source and target languages influence the BLEU scores of downstream tasks. To do these experiments, we used pre-trained MT models available on HuggingFace transformers. We used the following models: Spanish to Catalan, Catalan to Spanish, and English to French for Spanish to Catalan, Catalan to Spanish, Spanish to Portuguese, Portuguese to Spanish, French to Bambara and Bambara to French respectively. We computed the level of mutual intelligibility between the languages selected and the downstream tasks. 

\section{Related Work}\label{Lit-review}
In recent times, there has been an increase of research interest in  low-resource MT scenarios \cite{jawahar2021exploring, baziotis2020language}. NMT models, specifically those based on the Transformer architecture, have been shown to perform well when translating between similar languages \citep{przystupa2019neural, adebara2020translating, barrault2019findings, Barrault2020FindingsOT}, low resource scenarios \citep{adebara2021translating}, and in contexts not involving English~\cite{fan2021beyond}. 

Furthermore, pre-training techniques have been successful for many NLP tasks \cite{zoph2016transfer, durrani2021transfer} including NMT \cite{aji2020neural, weng2020acquiring}. Self-supervised pre-training acquires general knowledge from a large amount of unlabeled monolingual or multilingual data to improve the training process of downstream tasks~\cite{aji2020neural, devlin2018bert}. The pre-trained model acquires some syntactic and semantic knowledge which can be transferred as initialized parameters to improve NMT models and translation quality~\cite{goldberg2019assessing, jawahar2019does, aji2020neural}. The intuitive justification for using pre-trained models is that the embedding space becomes more consistent, with semantically similar words closer together.

The knowledge from pre-trained language models (LMs) can be used to initialize the NMT model before training it on parallel data. However, there are certain limitations for MT tasks. First, LMs cannot be easily fine-tuned for MT tasks. Second, there is a discrepancy between pre-training objectives for LMs and the training objective in MT. Existing pre-training approaches such as mBART rely on auto-encoding objectives to pre-train the models, which are different from MT. Furthermore, LMs learn to reconstruct all source tokens with some noises, while NMT learns to translate most source tokens and copy only a few of them. LM pre-training is said to copy about $65\%$ of tokens, while NMT training needs to copy less than $10\%$ \cite{knowles2018context}. The unexpected knowledge/bias can be therefore propagated to the NMT model via pre-training, which may result in NMT models mistakenly copying source tokens to the target side~\cite{liu2021copying}. For instance, because copying behaviours can be learned, a source word such as ``shoe" may be copied to the target by pre-training based NMT models instead of providing a translation. Therefore, fine-tuning MT models on pre-trained LMs still do not achieve adequate improvements. %In addition, existing MT pre-training approaches focus on using multilingual models to improve MT for low resource or medium resource languages. There has not been one pre-trained MT model that can improve for any pairs of languages, even for rich resource settings such as English-French.

In order to address the difference in training objectives that using pre-trained language models results in, we use pre-trained MT models to initialize our models. This is still a type of \textit{transfer learning}. 

Following the justification for pre-trained models, we hypothesize that two linguistically similar languages will share closer semantic and syntactic relationships. This is based on the assumption that the more similar the source and target languages, the more similar the syntax and semantic properties and the higher the gains from using pre-trained models will be. We now introduce our data.

\section{Data}\label{Data}
For our experiments, we use parallel data from OPUS~\cite{tiedemann-2012-parallel}. Our data are from the following language pairs Spanish and Catalan, Spanish and Portuguese, and French and Bambara. We use data in the two directions from each of these three pairs. Details about our data is in Table~\ref{tab:data}.

%\begin{table}[!ht]
%\begin{center}
%\textbf{}
%\small 
%\begin{tabular}{>r{2}r{2}r{2}} 
%\begin{tabular}{l|l|l|l|l}
% \toprule
%\multicolumn{1}{c}\textbf{Pair}  &\textbf{Lang} & \textbf{Sent} & \textbf{Words} \\ 
% \toprule

%\multicolumn{1}{c}{}  &es & $10$M & $284.6$M \\

%\multicolumn{1}{c}{\multirow{-2.5}{*}{\textbf{\small ES-CA}}}  &ca & $10$M & $273.3$M  \\

%\multicolumn{1}{c}{\multirow{-2}{*}{\rotatebox[origin=c]{90}{\textbf{\small ES-CA}}}}  &mono-es & $67.5$M & $1.7$B  \\
%\multicolumn{1}{c}{}   &mono-ca & $24.7$M & $733.9$M \\
%\hline%%%%%%%%%%%%%%%%%%%%%%%%%%%
 
%\multicolumn{1}{c}{}  &es & $4.1$M & $86.8$M \\
%\multicolumn{1}{c}{\multirow{-2.5}{*}{\textbf{\small ES-PT}}}   &pt & $4.1$M & $82.7$M  \\

%\multicolumn{1}{c}{\multirow{-2.2}{*}{\rotatebox[origin=c]{90}{\textbf{\small ES-PT}}}}  &mono-es & $67.9$M & $1.7$B  \\
%\multicolumn{1}{c}{}   &mono-pt & $13$M & $274.9$M \\
 %\hline%%%%%%%%%%%%%%%%%%%%%%%%%%%
 
% \multicolumn{1}{c}{}  & fr & $9.9$K & $179.3$M \\
%\multicolumn{1}{c}{\multirow{-2.5}{*}{\textbf{\small FR-BM}}}  & bm & $9.9$K  &  $202.9$M   \\

%\multicolumn{1}{c}{\multirow{-2.5}{*}{\rotatebox[origin=c]{90}{\textbf{\small FR-BM }}}}  &mono-fr &  -   &  -  \\
 %\multicolumn{1}{c}{}  &mono-bm &  - & - \\
% \toprule

%\end{tabular}
%\end{center}
%\caption{Number of sentences and words for the training data used for each language pair.}\label{tab:data}
%\end{table}

\begin{table}[!ht]
\small 
\centering
%\begin{adjustbox}{width=6cm}
%\renewcommand{\arraystretch}{1.5}{
\begin{tabular}{l|c|r|r}
 \toprule
\multicolumn{1}{c}{\textbf{Pair}} & \multicolumn{1}{c}{\textbf{Lang}} & \multicolumn{1}{c}{\textbf{Sent}} & \multicolumn{1}{c}{\textbf{Words}} % & %\multicolumn{1}{c}{\textbf{Untok TTR}} & \multicolumn{1}{c}{\textbf{Tok TTR}}
\\
 \toprule
\multirow{2}{*} {\textbf{ES-CA}}  & \textbf{ES} & $10$M & $284.6$M \\ %& $0.453$\% & $0.448$\%\\
                                  & \textbf{CA} & $10$M & $273.3$M \\
                        %& $0.44$\%  &$0.419$\%  \\
\multirow{2}{*} {\textbf{ES-PT}}   & \textbf{ES}  &$4.1$M  & $86.6$M \\%& $0.616$\% & $0.605$\% \\
                                  & \textbf{PT}  &$4.1$M  &$82.7$M   \\
                        %& $0.625$\% & $0.613$\% \\
\multirow{2}{*} {\textbf{FR-BM}}           & \textbf{FR}  &$9.9$K & $179.3$K  \\%& $10.03$\% & $ 9.22$\% \\
                                  & \textbf{BM}  &$9.9$K & $202.9$K \\  
                    %& $7.76$\% & $7.13$\%  \\
 \toprule
\end{tabular}%}
%\end{adjustbox}
\caption{Number of sentences and words for the training data used for each language pair. We also report the type token ratio (TTR) before and after tokenization.}\label{tab:data}
\end{table}

\subsection{Pre-Processing}
We perform pre-processing using the Moses toolkit~\citep{koehn2007moses}. For each language not supported by Moses, we use the tokenization setting of the language it is translated to. This applies only to Bambara, for which we used tokenization for the French language. We perform data cleaning, as we explain next. 

\section{Data Cleaning \& Analysis}\label{clean}
We perform data cleaning on the ES-CA, CA-ES, ES-PT, and PT-ES language pairs. We do not clean the French and Bambara pairs because we had very few training sentences for these. For cleaning, we run the langid tool~\cite{lui2012langid} on the concatenation of the source and target and remove sentences that are not identified as belonging to one or both of the language pair. In Table~\ref{tab:cleaned}, we provide some examples of data points we remove from the training data during data cleaning. These examples are removed because the claimed language is different from the language predicted by langid. After cleaning, we are left with $\sim 10$M clean sentences out of $\sim 18.3$M sentences for the Spanish and Catalan pair, and $\sim 3.1$M clean sentences out of $\sim 4.2$M sentences for the Spanish and Portuguese pair, respectively. We note that removed data comprise large portions of each dataset, thus confirming our concerns about data quality.  

\begin{table}[!ht]
\small
\begin{center}
\begin{tabular}{p{3cm}|p{1.5cm}|p{1.5cm}} 
 \toprule
\textbf{Sentence} & \textbf{Claimed} & \textbf{Predicted}\\
 \toprule

\textcolor{blue}{\textit{Animal Crossing:}} &  Spanish & English  \\ 

\textcolor{blue}{\textit{Pico de Santo Tomés}} &  Spanish & Portug. \\ 

\textcolor{blue}{\textit{Quinto Sereno Sammonico}} & Portug. & Italian \\ 

\textcolor{blue}{\textit{La sombra del caudillo}} &  Catalan & Spanish \\

\textcolor{blue}{\textit{Cultura del Nepal}} & Catalan & Spanish \\

\textcolor{blue}{\textit{Morts a Rāwalpindi}} & Catalan & French \\
 \toprule

 \end{tabular}
 \end{center}
 \caption{Examples removed from our training data. ``Claimed" refers to the expected language as coming from source, while ``predicted" is what langid.py identified. }\label{tab:cleaned}
 \end{table}

%Langid.py is an off-the-shelf language identification tool that comes with a pretrained model covering 97 languages \cite{lui2012langid}. Langid.py is trained with data from five different domains including government documents, software documentation, newswire, online encyclopedia and internet crawl. 

%\note[mam]{I did slight editing to the numbers above, but left original stats in a comment for our reference.}

% Original stats: After cleaning, we had $\sim 10,000,594$ clean sentences out of 18,278,450 sentences for the Spanish and Catalan pair, and 3,047,948 clean sentences out of 4,194,482 sentences for the Spanish and Portuguese pair respectively.

\section{Models}\label{model}
 %%%%%%%%%%%%%%%%%%%%%%%%%%%%%%%
 
\begin{table}[!ht]
\small
\begin{center}
\begin{tabular}{p{4cm}|p{2cm}} 
 \toprule
\textbf{Hyperparameter} & \textbf{Values} \\
 \toprule

encoder layers &  6 \\ 

decoder layers &  6 \\ 

attention heads & 8 \\ 

hidden layers &  6 \\

embedding dimension & 512 \\

dropout & 0.0 \\

vocab size & 49,621 \\
 \toprule

 \end{tabular}
 \end{center}
 \caption{Hyperparameter settings for the HuggingFace Marian Transformer models.}\label{tab:hyp}
 \end{table}

\begin{table}[!ht]
\begin{center}
\textbf{}
\small 
%\begin{tabular}{>r{2}r{2}r{2}} 
\begin{tabular}{>{}c|c|r|r}
 \toprule
\multicolumn{1}{c}{}  &\textbf{Model} & \textbf{\#Epochs} & \textbf{\#Highest} \\ 
 \toprule

\multicolumn{1}{c}{\multirow{-1}{*}{{\textbf{\small FR-BM}}}}  & tok & $100$ &  $55$\\
%%%%%%%%%%%%%%%%%%%%%%%%%%%

\multicolumn{1}{c}{\multirow{-1}{*}{{\textbf{\small BM-FR}}}}  & tok & $100$ &  $60$\\

\multicolumn{1}{c}{}  & untok & $6$ &  $3$\\
\multicolumn{1}{c}{\multirow{-2}{*}{{\textbf{\small ES-CA}}}}  & tok & $8$ & $3$ \\

\multicolumn{1}{c}{}  & untok & $7$ & $7$ \\
\multicolumn{1}{c}{\multirow{-2}{*}{{\textbf{\small CA-ES}}}}  & tok & $8$ & $8$ \\

\multicolumn{1}{c}{}  & untok & $17$ &  $15$\\
\multicolumn{1}{c}{\multirow{-2}{*}{{\textbf{\small ES-PT}}}}  & tok & $35$ &  $13$\\

\multicolumn{1}{c}{}  & untok & $18$ &  $16$\\
\multicolumn{1}{c}{\multirow{-2}{*}{{\textbf{\small PT-ES}}}}  & tok & $34$ & $23$ \\
%\multicolumn{1}{c}{\multirow{-2}{*}{{\textbf{\small PT-ES}}}}  & Baseline & $6$ & $6$ \\
 \toprule

\end{tabular}
\end{center}
\caption{Description the number of epochs for training each model and the epoch with the highest BLEU score.}\label{tab:epoch}
\end{table}
%%%%%%%%%%%%

\begin{table}[!ht]
\begin{center}
\small 
%\begin{tabular}{>r{2}r{2}r{2}} 
\begin{tabular}{>{}c|l|r|rH}
 \toprule
\multicolumn{1}{c}{}  &\textbf{Pair} & \textbf{Untokenized} & \textbf{Tokenized} & \textbf{Baseline} \\ 
 \toprule

\multicolumn{1}{c}{}  & es-ca & $77.3$ &  $86$ & -\\

\multicolumn{1}{c}{}  &ca-es & $87.7$ &  $87.8$ & -\\

\multicolumn{1}{c}{}  &es-pt & $46.9$ &  $47.3$ & -\\

\multicolumn{1}{c}{}  &pt-es & $52.9$ &  $53.6$ & $39.7$\\

\multicolumn{1}{c}{}  &fr-bm & -& $6.6$ &  -  \\

\multicolumn{1}{c}{}  &bm-fr & -& $6.07$ & -  \\

 \toprule
\end{tabular}
\end{center}
\caption{BLEU scores on our Dev set.}\label{tab:bleu}
\end{table}

%%%%%%%%%%%%%%%%%%%%%%%%%%%%%%%%%

%%%%%%%%%%%%%%%%%%%%%%%%%

\begin{table*}[!t]
\small 
\centering
\begin{adjustbox}{width=12cm}
\renewcommand{\arraystretch}{1.5}
{
\begin{tabular}{c|l|l|l|l|l|l|l}
\toprule
\multirow{2}{*}{\textbf{Pre-Trained Model}} & \multirow{2}{*}{\textbf{Pair}} & \multicolumn{3}{l}{\textbf{Untokenized System}} & \multicolumn{3}{l}{\textbf{Tokenized System}} \\

                       &                       & BLEU           & RIBES           & TER          & BLEU          & RIBES          & TER          \\
                       \toprule
\href{https://huggingface.co/Helsinki-NLP/opus-mt-ca-es}{ca-es}                 & ca-es                 & $76.8$            & $95.19$              & $15.421$           & $\textbf{82.79}$            & $\textbf{96.98}$             & $\textbf{10.918}$          \\
\href{https://huggingface.co/Helsinki-NLP/opus-mt-es-ca}{es-ca}                 & es-pt                 & $35.61$             & $82.48$             & \ $52.61$          & $38.10$           & $85.35$             & $46.556$ \\ 
\href{https://huggingface.co/Helsinki-NLP/opus-mt-ca-es}{ca-es}                 & pt-es                 & $43.86$             & $85.10$             & $43.801$          & $\textbf{47.71}$           & $\textbf{87.11}$             & $\textbf{39.213}$ \\ 
\href{https://huggingface.co/Helsinki-NLP/opus-mt-fr-en}{fr-en}                 & fr-bm                & -            & -            &-         & $1.32$           & $24.79$             & $97.89$ \\ 
\href{https://huggingface.co/Helsinki-NLP/opus-mt-fr-en}{en-fr}                 & bm-fr                & -            & -            &-         & $3.62$           & $36.17$             & $101.52$ \\ 
 \toprule
\end{tabular}}
\end{adjustbox}

\caption{BLEU, RIBES and TER Scores on the Test set for the Tokenized (primary) and untokenized (contrastive) configurations. The models for CA-ES and PT-ES language pairs were the best performing models highlighted in bold type.  }\label{tab:testbleu}
\end{table*}

%%%%%%%%%%%%%%%

\begin{table*}[!ht]
\centering
\begin{adjustbox}{width=16cm}
\renewcommand{\arraystretch}{1.5}{
\begin{tabular}{l|l|l}
\toprule
\multicolumn{1}{c}{\textbf{Pair}} & \multicolumn{1}{c}{\textbf{Category}} & \multicolumn{1}{c}{\textbf{Text}} \\
\toprule
\multirow{4}{*}{\textbf{ES-CA}}   & \textbf{Source} &  Por consiguiente, el Fondo debe movilizarse para aportar una contribución financiera en favor de Bulgaria, Grecia, Lituania y Polonia.  \\
                                  & \textbf{Untok Output}                 & \colorbox{blue!10}{Por lo tanto}, el \colorbox{blue!10}{Fondo debe movilizarse para que se conceda} una contribución financiera \colorbox{blue!10}{a} Bulgaria, Grecia, Lituania \colorbox{blue!10}{y} Polonia.                                  \\
                                  & \textbf{Tok Output}                   & Per tant, el Fons s'ha de mobilitzar per aportar una contribució financera \colorbox{red!10}{a} favor de Bulgaria, Grècia, Lituània i Polònia.                              \\
                                  & \textbf{Reference}  & 
                                  Per tant, el Fons s’ha de mobilitzar per aportar una contribució financera en favor de Bulgària, Grècia, Lituània i Polònia.
                                  \\
                                  \toprule
\multirow{4}{*}{\textbf{CA-ES}}   & \textbf{Source Text} &  A fi de reduir al mínim el temps necessari per mobilitzar el Fons, aquesta Decisió s’ha d’aplicar a partir de la data de la seva adopció,  \\
                                  & \textbf{Untok Output} & \colorbox{blue!10}{A} fin de reducir al mínimo el tiempo necesario para movilizar el Fondo, \colorbox{blue!10}{esta} Decisión debe aplicarse a partir de la fecha de su adopción,                                \\
                                  & \textbf{Tok Output}  &  Con el fin de reducir al mínimo el tiempo necesario para movilizar el Fondo, \colorbox{red!10}{esta} Decisión \colorbox{red!10}{se ha de aplicar} a partir de la fecha de su adopción.                           \\
                                  & \textbf{Reference}  & 
                                    Con el fin de reducir al mínimo el tiempo necesario para movilizar el Fondo, la presente Decisión debe aplicarse a partir de la fecha de su adopción,
                                  \\
                                  \toprule
\multirow{4}{*}{\textbf{ES-PT}}   & \textbf{Source Text} &  Posición del Parlamento Europeo de 6 de abril de 2017 (pendiente de publicación en el Diario Oficial) y Decisión del Consejo de 11 de mayo de 2017. \\
                                  & \textbf{Untok Output} & Posição do Parlamento Europeu de 6 de \colorbox{blue!10}{A}bril de 2017 (\colorbox{blue!10}{pendente de publicação} no Jornal Oficial) e \colorbox{blue!10}{D}ecisão do Conselho de 11 de Maio de 2017.\\
                                  & \textbf{Tok Output}  &  Posição do Parlamento Europeu de 6 de \colorbox{red!10}{A}bril de 2017 (\colorbox{red!10}{indiferente à publicação} no Jornal Oficial \colorbox{red!10}{da União Europeia} e decisão do Conselho de 11 de Maio de 201 ).\\
                                  & \textbf{Reference}  & Posição do Parlamento Europeu de 6 de abril de 2017 (ainda não publicada no Jornal Oficial) e decisão do Conselho de 11 de maio de 2017.\\
                                  \toprule  
\multirow{4}{*}{\textbf{PT-ES}}   & \textbf{Source Text} & Os Estados-Membros transmitem os dados referentes ao transporte por vias navegáveis interiores no seu território nacional à Comissão (Eurostat). \\
                                  & \textbf{Untok Output} & Los Estados miembros transmitirán \colorbox{blue!10}{a la Comisión (Eurostat) } los datos relativos al transporte por vías navegables interiores en su territorio nacional. \\
                                  & \textbf{Tok Output}  &  Los Estados miembros transmitirán \colorbox{red!10}{a la Comisión} los datos relativos al transporte por vías navegables interiores en su territorio nacional (convenientREAT)\\
                                  & \textbf{Reference}  & Los Estados miembros transmitirán los datos relativos al transporte por vías navegables interiores en su territorio nacional a la Comisión (Eurostat).\\
                                  \toprule
\multirow{4}{*}{\textbf{FR-BM}}   & \textbf{Source Text} & vous pourriez peut-être organiser de petits groupes pour lire et discuter de ce livre, chapitre par chapitre.\\
                                  & \textbf{Tok Output}  & \colorbox{blue!10}{- An y'a m\textepsilon n kura in m\textepsilon n: Mama denmuso, an ka jamana de wa k'a furu min ma k\textopeno n\textopeno.}\\
                                  & \textbf{Reference}  & aw b\textepsilon se ka to ka m\textopeno g\textopeno w dalaj\textepsilon ka gafe in kalan; ani ka hakilina falenfalen k\textepsilon sigidaw kan kelen kelen.\\
                                  \toprule   
\multirow{4}{*}{\textbf{BM-FR}}   & \textbf{Source Text} & Hakilijigin ka \textltailn\textepsilon sin k\textepsilon n\textepsilon ya­baarak\textepsilon law ma\\
                                  & \textbf{Tok Output}  & cher agent de santé villageois,\\
                                  & \textbf{Reference}  & cher agent de santé villageois,\\
                                  \toprule                            
\end{tabular}}
\end{adjustbox}
    \caption{Examples sentences from the various pairs and corresponding translations based on the untokenized and tokenized models. Examples are from the Dev set. We highlight the differences between the outputs from the untokenized model and the reference text with~\colorbox{blue!10}{blue highlights} and the differences between the tokenized model and the reference text in~\colorbox{red!10}{red highlights}. It can be observed that the number of errors in the untokenized model (based simply on the number of blue highlights here) is larger than that in the tokenized model (less errors/red highlights}
    
    \label{tab:exp}
\end{table*}

%%%%%%%%%%%%%%%%%%%%
\subsection{Primary and Contrastive Models}
We developed our \textit{primary} and \textit{contrastive} models using Transformers from HuggingFace library ~\cite{wolf2019huggingface}. The primary models were developed using tokenized data while the contrastive models employed untokenized data. For the tokenized setting, we used Moses tokenizer (as explained earlier) while the untokenized setting used the data just as they were made available to us by shared task organizers. 

We used the pre-trained NMT models developed by Helsinki-NLP on HuggingFace. We used pre-trained models closest to the language pairs we trained. For language pairs without existing pre-trained models, we used a close language pair with either the source or target matching one of our downstream task languages in a given pair. Specifically, we used the following Marian models released by Helsinki-NLP: \href{https://huggingface.co/Helsinki-NLP/opus-mt-es-ca}{es-ca} (for ES-PT),~\href{https://huggingface.co/Helsinki-NLP/opus-mt-ca-es}{ca-es} (for CA-ES and PT-ES), ~\href{https://huggingface.co/Helsinki-NLP/opus-mt-en-fr}{fr-en} (for FR-BM), and~\href{https://huggingface.co/Helsinki-NLP/opus-mt-en-fr}{en-fr} (for BM-FR).

As an example, we report the hyperparameters for the CA-ES \textit{primary} model in Table~\ref{tab:hyp}. This model had the best BLEU and RIBES score for this language pair. We trained each model for different number of epochs due to time and GPU constraints. We show the number of epochs each model is trained for and the epoch with the highest BLEU score in Table~\ref{tab:epoch}. We did not train any \textit{contrastive} models for FR-BM and BM-FR pairs, so we report the training for the primary (tokenized) models only. 

\subsection{Baseline}

We developed a single baseline model based on Transformers as implemented in Fairseq. This model does not use any pre-trained MT models nor tokenization. This model was developed for the ES-PT pair for six epochs and it achieved a BLEU score of $37.6$. For comparison, we developed a model for the same pair (i.e., ES-PT) based on an already available pt-es pre-trained MT model. After six epochs, this ES-PT model employing transfer learning achieved $52.18$ BLEU (thus significantly outperforming our baseline). Based on this result, we resumed with experiments for all other language pairs \textit{without} including a baseline. Ideally, we would train such baseline models for all the pairs. However, due to limited time and GPU resources, we only trained a baseline for a single pair.

%%%%%%%%%%%%%%%%%%%%
\section{Evaluation} \label{Evaluation}
We evaluated our models on both the Dev and Test sets. We used the checkpoint with the best BLEU score as evaluated on DEV as our best model. We used a beam size of four during evaluation on both Dev and Test and evaluated on de-tokenized data.

%\subsection{Evaluation on Dev set}
\textbf{Evaluation on Dev set.} We report the results on the Dev sets for each language pair in Table~\ref{tab:bleu}. As explained, the models were trained with both tokenized and untokenized data. As Table~\ref{tab:bleu} shows, the tokenized setting yielded the highest performance for \textit{all} language pairs. We show sample outputs from our tokenized models (from Dev data) in Table~\ref{tab:exp}.

%\subsection{Evaluation on Test set}
\textbf{Evaluation on Test set.} Our Test set performance was evaluated by the shared task organizers using BLEU ~\cite{papineni2002bleu}, RIBES, and TER ~\cite{snover2009fluency}. We report the scores in Table~\ref{tab:testbleu}. Each of our CA-ES and PT-ES models ranked top 1 based on the official shared task results. In addition, we were the only team to submit models in the official competition for the French-Bambara pairs. As with the Dev set, the tokenized setting gave the highest performance for all language pairs.

\section{Effect of Language Similarity}\label{effect}
%We wanted to quantify the effect that the languages used for pretraining had on the downstream tasks. To determine the effect of language similarity on NMT pretraining, we computed the.

In order to gain some insight into the interference of similarity between languages of a given pair, we performed an analysis based on Levenstein distance that allows us to identify the percentage of cognates shared between the languages. We then compared system output to the reference sentences, trying to quantify how much the system is able to translate cognates correctly (in this case the correct translation will have the same cognate word in the target as it is in the source). We performed this analysis for one language pair: CA-ES and found that the model learned the cognates correctly up to ~80\% of the time.

% We investigate the type of errors that occur with similar language translation when fine-tuned on pretrained MT models. Specifically we wanted to investigate whether the errors occurred with cognate words or other words. To do these experiments, we used pre-trained MT models available on HuggingFace transformers. We used ES-CA pre-trained model to finetune both ES-CA and ES-PT language pairs; we used CA-ES pre-trained model to finetune CA-ES and PT-ES language pairs and finally the FR-BM and BM-FR models were finetuned on ES-FR pre-trained models.

%\note[mam]{I suggest talking about different types of models and the different settings. What tools did we use, what hyperparameters (perhaps put these in a table), and how many epochs each model took to converge. Also explain the tokenization/no-tokenization, and the eval metrics we report in. Try to discuss the results and explain why do you think something works. If you are able to report the total number of parameters in each model, that will also be nice. It will also be nice to include some example translations from the best model. If you have time, do a quick error analysis perhaps on 2 pairs, but I realize you may not be able to do that, so it's ok.}

\section{Conclusion}\label{Conclusion}
We describe our contribution to the WMT2021  Similar Languages Translation Shared Task. We develop models for ES-CA, CA-ES, ES-PT, PT-ES, FR-BM, and BM-FR and show the improvement our models make with tokenized data when compared to untokenized data. We also show the utility of transfer learning based on fine-tuning NMT pre-trained models. Future work can investigate how the choice of pre-trained models affects the downstream tasks. 
 \section*{Acknowledgements}
 We gratefully acknowledge support from the Natural Sciences and Engineering Research Council of Canada (NSERC), the Social Sciences Research Council of Canada (SSHRC), Compute Canada (\url{www.computecanada.ca}), and UBC ARC--Sockeye (\url{https://doi.org/10.14288/SOCKEYE}).
 
% \note[mam]{I suggest you talk to Moatez or Chiyu about how to add URLs to the references. It will also be nice if we extend the biliography, citing new works as well as our own previous work on similar languages. Also work on low-resource (including Michael's), and code-switched MT (including Moatez, Abdu and Ganesh's). Perhaps you can say recently there is a growing body of research on similar languages (e.g., xxx), low-resource MT (e.g., xxx) including on Indigenous languages (e.g., xxx), code-switched MT (e.g., xxx).}
\normalem
\bibliography{anthology,emnlp2020}
\bibliographystyle{acl_natbib}

\end{document}